\documentclass{article}

\usepackage[final]{corl_2020} % Uncomment for the camera-ready ``final'' version.

\usepackage{graphics} % for pdf, bitmapped graphics files
\graphicspath{ {/img/} }
\usepackage{epsfig} % for postscript graphics files
\usepackage{algorithm}
\usepackage{algorithmic}
\usepackage{amsmath,amssymb}  
\usepackage{caption}
\usepackage{mathtools}  
\usepackage{hyperref}
\usepackage{booktabs}
\usepackage{siunitx}
\usepackage{graphics}
\usepackage{epstopdf} 
\usepackage{xcolor}
\usepackage{fontenc}
\usepackage{comment}
\usepackage{multirow}
\usepackage{multicol}
\usepackage{graphicx}
\usepackage[nolist]{acronym}
\usepackage{calrsfs}
\usepackage[capitalise]{cleveref}
\usepackage{placeins}
\usepackage{subcaption}
\usepackage{tabularx}
\usepackage{diagbox}
\usepackage{verbatim}
\usepackage{color}
\usepackage{xcolor}
\usepackage[normalem]{ulem}
\usepackage{balance}
\usepackage{wrapfig,lipsum}
\usepackage{dirtytalk}
\DeclareMathAlphabet{\pazocal}{OMS}{zplm}{m}{n}

\DeclareMathOperator*{\argmin}{arg\,min}

\newcommand{\BF}[1]{{\color{blue}{{#1}}}} % Bruno questions
\title{Social-VRNN: One-Shot Multi-modal Trajectory Prediction for Interacting Pedestrians}

\author{
	Bruno Brito\\
	Cognitive Robotics Department\\
	Delft University of Technology \\
	Netherlands\\
	\texttt{bruno.debrito@tudelft.nl} \\
	\And
	Hai Zhu \\
	Cognitive Robotics Department\\
	Delft University of Technology \\
	Netherlands \\
	\texttt{h.zhu@tudelft.nl} \\
	\AND
  	Wei Pan \\
	  Cognitive Robotics Department\\
	  Delft University of Technology \\
  	Netherlands \\
   	\texttt{wei.pan@tudelft.nl} \\
  	\And
  	Javier Alonso-Mora \\
	  Cognitive Robotics Department\\
	  Delft University of Technology \\
  	Netherlands \\
	\texttt{j.alonsomora@tudelft.nl} \\
}

\begin{document}
\maketitle

%===============================================================================

\begin{abstract}
	Prediction of human motions is key for safe navigation of autonomous robots among humans. In cluttered environments, several motion hypotheses may exist for a pedestrian, due to its interactions with the environment and other pedestrians.
	Previous works for estimating multiple motion hypotheses require a large number of samples which limits their applicability in real-time motion planning. In this paper, we present a variational learning approach for interaction-aware and multi-modal trajectory prediction based on deep generative neural networks. 
	Our approach can achieve faster convergence and requires significantly fewer samples comparing to state-of-the-art methods. Experimental results on real and simulation data show that our model can effectively learn to infer different trajectories. We compare our method with three baseline approaches and present performance results demonstrating that our generative model can achieve higher accuracy for trajectory prediction by producing diverse trajectories. 
\end{abstract}

\vspace{-0.4cm}
\suppmaterial{\url{https://youtu.be/tBr5v7TXyG0}}
\vspace{-0.4cm}
\code{\url{https://github.com/tud-amr/social_vrnn.git}}
%===============================================================================

\section{Introduction}
	
%WHat is the problem
%\wei{I will rewrite your intro: One of the key features required in autonomous navigation systems to successfully navigate in crowded environments is the ability to predict the other agent's behavior effectively. The successful introduction of mobile robots in human environments is strongly dependent on their ability to reason about the other agent intentions and to forecast their motions. Such skill will allow state-of-art motion planning algorithms to generate safe and socially compliant motion plans for a vast number of applications ranging from service robots to autonomous driving. }
Prediction of human motions is key for safe navigation of autonomous robots among humans in cluttered environments.
%To predict pedestrians behavior is the key for efficient autonomous navigation in clutter environment.
Therefore, autonomous robots, such as service robots or autonomous cars, shall be capable of reasoning about the intentions of pedestrians to accurately forecast their motions. Such abilities will allow planning algorithms to generate safe and socially compliant motion plans \cite{8768044,Zhu2020ICRA}. 
% Especially in the scenario of interaction with human, the autonomous robots should be able to predict human motions in order for safe navigation.

%in such environments even with perfect predictions, the state-of-art motion planners may fail under these circumstances due to the Freezing Robot Problem (FRP) if the interaction between the agents is not taken into account \cite{Trautman2010}. The reason is the lack of cooperation which makes the robot believe that no possible path might exist. 
%Why is it interesting/important? E.g. possible applications
%The successful introduction of mobile robots in human environments is strongly dependent on their ability to reason about the other agent intentions and to forecast their motions. \BF{I'll remove this paragraph if necessary due to space limitations}
%Such skill will allow state-of-art motion planning algorithms to generate safe and socially compliant motion plans for a vast number of applications ranging from service robots to autonomous driving. 

%For instance, humans predict the state of environment several steps ahead and plan their motions accordingly.

%to a set of social acceptable rules.

%However, it is intractable to fully enumerate and program all these rules and scenarios into an algorithm to predict their motions.
%freezing robot problem

%\BF{Why do we need the multimodal prediction?}

\begin{wrapfigure}{r}{.5\textwidth}
	\centering
	\includegraphics[trim=0 0 0 40,width=.5\textwidth]{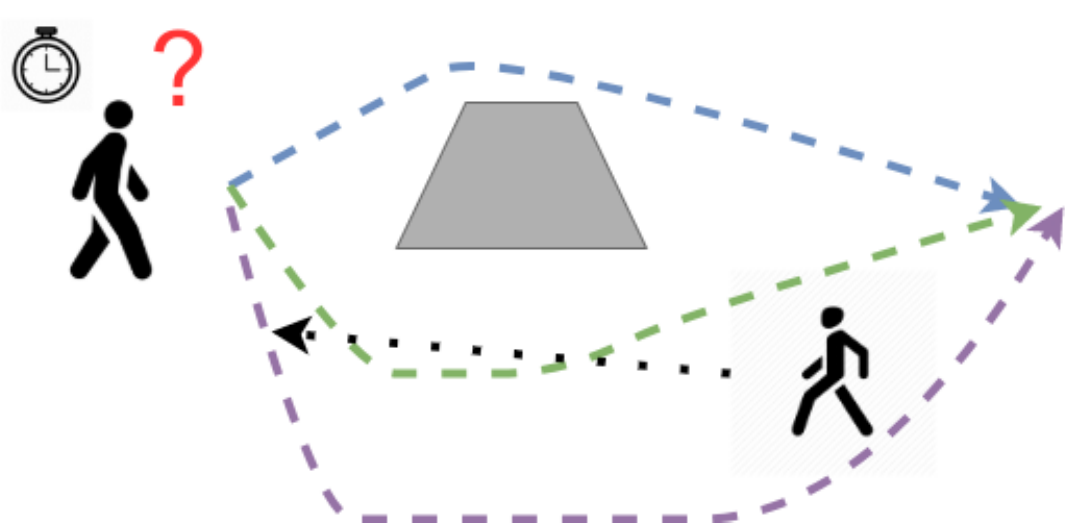}
	\caption{Illustration of a scenario where there are multiple ways that
	two pedestrians can avoid a collision. We present a method
	that given the same observed past, predicts multiple socially
	acceptable trajectories in crowded scenes.}\label{fig:multiple_path_example}
\end{wrapfigure}

Generally, human motions are inherently uncertain and multi-modal \cite{kothari2020human}.
% Looking at pedestrians motion, the human motion is inherently uncertain and multi-modal \cite{kothari2020human}.
%Why is it hard / challenging?
The uncertainty is caused by partial observation of the pedestrians' states and their stochastic dynamics. The multimodality is due to interaction effects between the pedestrians, the static environment and non-convexity of the problem. 
For instance, as Fig. \ref{fig:multiple_path_example} shows, a pedestrian can decide to either avoid a static obstacle or engage in a non-verbal joint collision-avoidance maneuver with the other upcoming pedestrian, avoiding on the right or left. Hence, to accurately predict human motions, inference models providing multi-modal predictions are required.

%For instance, as depicted in Fig. \ref{fig:multiple_path_example}, under the same circumstances a pedestrian may follow a different trajectory either to avoid a static obstacle or to engage in a non-verbal joint collision avoidance with the other pedestrian.

%Predicting the future trajectories of agents navigating in an unconstrained crowded environment is challenging due to the different factors influencing their motion and its properties:

%\begin{enumerate}
%    \item \textbf{Interaction}: The motion of each agent influences each other. For instance, humans cooperate to avoid collisions.
%    \item \textbf{Static environment}: The static obstacles surrounding the agent imposes physical constraints on its motion. 
%    \item \textbf{Dynamics}: Each agent has a intrinsic motion model which is kinodynamically constrained.
%    \item \textbf{Stochastic}: Trajectories vary from run to run. For instance the human motion is inherently stochastic. 
%    \item \textbf{Multimodal}: Under the same situation an agent may follow different paths which are also socially compliant.
%\end{enumerate}

%E.g. why do naive approaches fail? Or, what's wrong with previous n solutions?
%\BF{I agree that this paragraph is more RL section but I am following Dariu' s recommend structure that says: E.g. why do naive approaches fail? Or, what's wrong with previous n solutions?}
A large number of prediction models have been proposed. However, some of these approaches only predict the mean behavior of the agents \cite{Pfeiffer2017}. Others apply different techniques to model uncertainty such as ensemble modeling \cite{lotjens2019safe}, dropout during inference \cite{gal2016dropout} or learn a generative model and generate several trajectories by sampling randomly from the latent space \cite{Gupta2018}. Recently, Generative Adversarial Networks (GANs) have been employed for multi-modal trajectory prediction by randomly sampling the latent space to generate diverse trajectories \cite{amirian2019social}. Nevertheless, these methods have two main drawbacks. First, GANs are difficult to train and may fail to converge during training. Second, they require a large number of samples to achieve good prediction performance which is impracticable for real-time motion planning. Moreover, these approaches assume an independent prior across different timesteps ignoring the existing time dependencies on trajectory prediction problems.

%ensemble modelling is limited to really structured environment, such as driving scenarios, where the different hypothesis can be known a priori.
%\begin{figure}[t]
%    \hfill\begin{minipage}{.5\textwidth}\centering
%		\includegraphics[scale=0.35]{img/multiple_path.pdf}
%		\caption{Illustration of a scenario where there are multiple ways that
%		two pedestrians can avoid a collision. We present a method
%		that given the same observed past, predicts multiple socially
%		acceptable trajectories in crowded scenes.}
%	\end{minipage}
%   \label{fig:multiple_path_example}
%\end{figure}

The objective of this work is to develop a prediction model suitable for interaction-aware autonomous navigation.
% In this paper, we aim to develop a prediction model suitable for autonomous navigation.
Hence, we  address  these  limitations with a novel generative model for multi-modal trajectory prediction based on Variational Recurrent Neural Networks (VRNNs) \cite{vrnn}. We treat the multi-modal trajectory prediction problem as modeling the joint probability distribution over sequences. 

%\BF{this is more method overview: move the beginning of the third sentence}
%To this end, our model first creates a joint representation of three input channels: the pedestrian dynamics, the static environment and other interacting pedestrians. Then, the VRNN structure learns the complex multi-modal distribution conditional to the previous timesteps. Finally, we use a Gaussian Mixture Model (GMM) as output probability model enabling one-shot multi-modal trajectory predictions.
%encoding the uncertainty and multimodality of the data.
%The state-of-art approaches have at least one of the following drawbacks: (i) rely on hand-crafted functions \cite{Helbing1995} - \cite{Kretzschmar2016}
%(ii) do not not reason about static obstacles such
%as the workspace limits or other static obstacles of the robot
%environment \cite{alahi2016social}.

%The existing approaches are limited by at least one of the following shortcomings: (i) The feature functions, which abstract agent trajectory information to an internal representation, are hand-crafted and therefore can only capture simple interactions. (ii) The approaches are not scalable to dense crowds since they use pairwise interactions between all agents [1]–[3], which leads to a quadratic complexity in number of agents, and therefore real-time computation is only feasible for a small number of agents. (iii Static obstacles are neglected [1], [4], [5] and (iv) knowledge about a set of potential destinations is assumed [1], [3], [5].
This paper's main contribution is a new interaction-aware variational recurrent neural network (Social-VRNN) design for one-shot multi-modal trajectory prediction. By following a variational approach, our method achieves faster convergence in comparison with GAN -based approach. Moreover, employing a time-dependent prior over the latent space enables our model to achieve state-of-the-art performance and generate diverse trajectories with a single network query.
%\BF{Improve contributions. remove paragraph. the main contribution is the first one and then in order to achieve that we use that training strategy. what is the advantage over  state-of-art.}
%\begin{itemize}
%    \item A new interaction-aware variational recurrent neural network(Social-VRNN) design for one-shot multi-modal trajectory prediction;
%    \item A training strategy is proposed to learn more diverse trajectories in an interpretable fashion;%An efficient sampling approach is proposed by modelling social and environment context as \emph{prior};
   % A sampling approach to generate trajectories from different modes of the probability space accounting for the social and environment context
    %\item  %Highly discriminated trajectories can be predicted compared to state-of-the-art approaches, as demonstrated on both simulated and real datasets;
    %Experimental results on real and simulation data demonstrating the ability of our method to produce diverse trajectories%A simulated dataset to specifically evaluate the multimodality of predicted distributions 
%\end{itemize}

To this end, we propose a training strategy to learn more diverse trajectories in an interpretable fashion. Finally, we present experimental results demonstrating that our method outperforms the state-of-the-art methods on both simulated and real datasets using one-shot predictions.

%===============================================================================

\section{Related Works}\label{sec:related_work}

%FRP happens when a crowd is so dense that the robot is not able to find any safe path, causing it to stop any motion.
%Approaches that do not incorporate the interactions between agent's during the path planning task fail under crowed environments. 
%The reason is the lack of cooperation which makes the robot believe that no possible path might exist. 

Early works on human motion prediction are typically model-based. In \cite{Helbing1995}, a model of human-human interactions was proposed by simulating attractive and repulsive physical forces denominated as \say{social forces}.
%\wei{plz use passive vocie like I re-write in the above sentence when you say what other work do. There are quite a lot in this section, please address}
To account for human-robot interaction, a Bayesian model based on agent-based velocity space was proposed in \cite{Kim2015}. However, these approaches do not capture the multi-hypothesis behavior of the human motion. To accomplish that, \cite{Trautman2010} proposed a path prediction model based on Gaussian Processes, known as interactive Gaussian Processes (IGP). This was done by modeling each individual's path with a Gaussian Process. The main drawbacks of this approach are the usage of hand-crafted functions to model interaction, limiting their ability to learn beyond the perceptible effects, and is computationally expensive.

Recently, Recurrent Neural Networks (RNNs) have been employed in trajectory prediction problems \cite{Becker2018}.
Building on RNNs, a hierarchical architecture was proposed in \cite{Xue2018} and \cite{Pfeiffer2017}, which incorporated information about the surrounding environment and other agents, and performed better than previous models.
%\cite{Xue2018} presents a hierarchical LSTM-based encoder-decoder network combining information about the environment, other agents and the ego-agent. Similarly, \cite{Pfeiffer2017} proposes to predict the agent's velocities instead of its positions. 
Despite the high prediction accuracy demonstrated by these models, they are only able to predict the average behavior of the pedestrians.

In contrast, Social LSTM \cite{alahi2016social} models the prediction state as a Bivariate Gaussian and thus, uncertainty can be incorporated. Moreover, interaction is modeled by changing the hidden state of each agent network according to the distance between the agents, a mechanism know as "Social pooling".
%Moreover, to incorporate interaction \cite{alahi2016social}, proposed to use a Long Short-Term Memory (LSTM) network to encode each agent trajectory and change the hidden state of each agent network according with the distance between the agents, a mechanism know as "Social pooling". 
%Then, to model the interaction effects of the other agents,  combined with a "Social" pooling mechanism to model interaction. This grid based polling mechanism changes the hidden state of each agent network according with the distance between the agents, based on the assumption near neighbors influence more each other than far neighbors.
%deep learning has gained increased attention to solve the motion prediction problem due to their ability to encode an huge amount of past information. On the latter, several works proposed different neural networks architectures to predict pedestrian trajectories based on observed tracklets. \cite{Becker2018} analyzed some of the proposed architectures, concluding that Recurrent Neural Networks (RNNs)with a Dense layer stacked on top achieved the best prediction performance. 
Several approaches extended the latter either by incorporating other sources of information or proposing updates in the model architecture improving the performance of the model. For instance, head pose information from the other agents was incorporated in \cite{hasan2018mx} resulting in a significant increase of the prediction accuracy. Context information from visual images was used to encode both human-human and human-space interactions \cite{bartoli2018context}. %context-aware probabilistic model: “context-aware” pooling which is able to learn and encode both human-human and human space interactions
Social pooling has been extended to generate collision-free predictions \cite{xu2018collision} and to preserve spatial information by employing Grid LSTMs \cite{lerner2007crowds}.

However, previous approaches did not consider the inherent multi-modal nature of human motions. In \cite{Gupta2018}, a generative model based on Generative Adversarial Networks (GANs) was developed to generate multi-modal predictions by randomly sampling from the latent space. This approach was extended with two attention mechanisms to incorporate information from scene context and social interactions \cite{Sadeghian}. However, GANs are very susceptible to mode collapsing causing these models to generate very similar trajectories. To avoid mode collapse, a recently improved Info-GAN for multi-modal trajectory prediction was proposed \cite{amirian2019social}. Besides, \cite{makansi2019overcoming} proposed a different training strategy to overcome the latter issue and improve trajectory prediction diversity. To account for the environment constraints, \cite{zhao2019multi} proposed to include scene context information provided by a top-view camera of the scene. However, such information is not available in a real autonomous navigation scenario. Moreover, to improve social interaction modelling, Graph Neural Networks have been used in \cite{chandra2020forecasting,eiffert2020probabilistic}.
Nonetheless, GANs are very difficult to train and typically require a large number of iterations until it converges to a stable Nash equilibrium. %Moreover, the environment context was not incorporated in \cite{amirian2019social}, and the predicted trajectories may collide with static obstacles. %To overcome the later issue, \cite{zhao2019multi} proposed to include scene context information provided by a top-view camera of the scene, making use of spatial relationships among agents and had shown good results in the Zara dataset. However, such information is not available in a real autonomous navigation scenario. 

Similar to our approach, the \textit{Trajectron++} \cite{salzmann2020trajectron++} employs variational learning to improve training convergence and speed. Kernel-based methods employed Mixture Density Networks (MDNs) to build a continuous map capturing the possible motion directions \cite{zhi2019spatiotemporal} or to learn a multi-model distribution over a set of trajectories \cite{zhi2020kernel}. Nevertheless, \cite{trajectron} assumes a time-independent prior over the latent space, and \cite{zhi2019spatiotemporal,zhi2020kernel} requires a large number of samples to produce distinct trajectories. In contrast, we propose a novel architecture to learn a multi-modal prediction model based on VRNNs that can significantly improve the network prediction performance and diversity. Moreover, our method only uses local information enabling its application for autonomous navigation.

\section{Variational Recurrent Neural Network}\label{sec:approach}
%\wei{I change the section title. In general, this section is well written.}
%\wei{throughout the paper, change "model" to "deep neural networks" or "deep net" or "network"}

\begin{figure*}[!t]
    \centering
    \includegraphics[width=\textwidth]{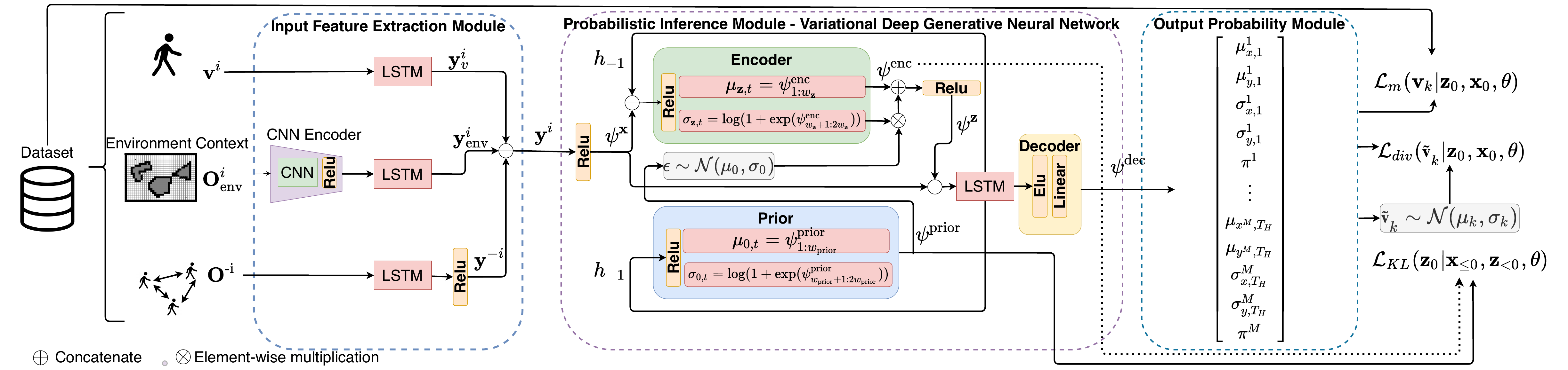}
    \caption{VRNN architecture for multi-modal trajectory prediction composed by: an input feature extraction, a probabilistic inference and output probability module. The first creates a joint representation of the input data $\mathbf{y}^i_{}=\{\mathbf{y}^{i}_{\mathbf{v}},\mathbf{y}^{i}_{\textrm{env}},\mathbf{y}^{-i}_{}\}$. The probabilistic inference module (Section \ref{sec:stochectic_module}) is based on the VRNN \cite{vrnn} incorporating: a encoder network to approximate a time-dependent posterior distribution $q(\mathbf{z}_{0}|\mathbf{x}_{\le 0},\mathbf{z}_{<0}) \sim \mathcal{N}(\mathbf{\mu}_{\mathbf{z},0},\textrm{diag}(\mathbf{\sigma}^2_{\mathbf{z},0}))$ with 
    $[\mathbf{\mu}_{\mathbf{z},0},\mathbf{\sigma}_{\mathbf{z},0}]=\psi^{\textrm{enc}}(\psi^{\mathbf{x}}(\mathbf{x}_0),\mathbf{h}_{-1},\theta_{q})$ with $\theta_{q}$ as the approximate posterior model parameters; a decoder network to model the conditional generation distribution $\mathbf{\mathbf{v}}_{k} | \mathbf{x}_{0}, \mathbf{z}_{0} \sim \mathcal{N}(\mathbf{\mu}_{\mathbf{x},0},\textrm{diag}(\mathbf{\sigma}^2_{\mathbf{x},0}))$ with $
    [\mathbf{\mu}_{\mathbf{v},1:T_H},\mathbf{\sigma}_{\mathbf{v},1:T_H}]=\psi^{\textrm{dec}}(\psi^{\mathbf{z}}(\mathbf{z}_0),\psi^{\mathbf{x}}_t(\mathbf{x}_0),\mathbf{h}_{-1},\theta_{\mathrm{dec}})$ with $\theta_{\mathrm{dec}}$ as the inference model parameters; a prior on the latent random variable $\mathbf{z} \sim \mathcal{N}(\mu_{\textrm{prior},0},\mathbf{\sigma}_{\textrm{prior},0})$ conditional to the hidden-state of the decoder network $[\mathbf{\mu}_{\textrm{prior},0},\mathbf{\sigma}_{\textrm{prior},0}]=\psi^{\textrm{prior}}(\mathbf{h}_{-1},\theta_{\textrm{prior}})$ with parameters $\theta_{\mathrm{prior}}$. Finally, the output probability module is a GMM (Section \ref{sec:output_model}). %One can refer to Section \ref{sec:approach} for more details.
    }%
    \label{fig:network_architecture}%
\end{figure*}

In this section, we present our Variational Recurrent Neural Network (VRNN) for multi-modal trajectory prediction, depicted in Fig. \ref{fig:network_architecture}. %To this end, our model first employs a feature extractor module (Section \ref{sec:inputs}) to create a joint representation of three input channels: the pedestrian dynamics, the static environment and other interacting pedestrians. Then,  the probabilistic inference module (Section \ref{sec:stochectic_module}) learns the complex multi-modal distribution conditional to the previous timesteps. Finally, the probability output module (Section \ref{sec:output_model}) applies a Gaussian Mixture Model (GMM) as output probability model enabling one-shot multi-modal trajectory predictions.

%We start with the problem formulation of multi-modal trajectory prediction. 
%Notations are and the notation used along with the article (Section \ref{sec:problem}).
%Then, we describe the first module which creates the input representation that is used to compute the multimodal predictions (Section \ref{sec:inputs}) followed by the network model used to learn the underlying a multimodal probability distribution (Section \ref{sec:stochectic_module}) and output distribution model used (Section \ref{sec:output_model}). 
%Later in Section \ref{sec:sampling}, we introduce a training strategy which allows to generate different modal trajectories by incorporating the social and environment context of the \textit{query-agent}. %sampling mechanism which allows to generate different modal trajectories by incorporating the social and environment context of the query-agent. 
%Finally, we define the loss function and explain the training procedure in Section \ref{sec:train}.

%model loss which motivates the model to learn different hypothesis and thus,solving the mode collapse issue.
\begin{comment}
\end{comment}
\subsection{Multi-modal Trajectory Prediction Problem Formulation}\label{sec:problem}

%This paper presents a data-driven approach to predict the uncertain and multimodal pedestrian navigation behavior.
Consider a navigation scenario with $n$ interacting agents (pedestrians) navigating on a plane $\pazocal{W} = \mathbb{R}^2$. The dataset $\mathbf{D}$ contains information about the $i$-th pedestrian trajectory $\tau^i_{1:N}=\{(\mathbf{p}_1^i,\mathbf{v}_1^i),\dots,(\mathbf{p}_N^i,\mathbf{v}_N^i)\}$ with $N$ utterances and their corresponding surrounding static environment $\pazocal{O}^{i}_{\textrm{env}} \subset \pazocal{W}$, for $i \in [0, \dots n]$. 
%$p^i_t$ is the agent position and $v_t^i$ the agent velocity at time $t$ in world coordinates. 
$\mathbf{v}^{i}_t = \{v^i_{x,t},v^i_{y,t} \}$ is the velocity and $\textbf{p}^{i}_{t} = \{p^i_{x,t},p^i_{y,t} \}$ is the position of the $i$-th pedestrian at time $t$ in the world frame. %Let $\mathbf{x}^{i}_t=\mathbf{v}^{i}_t$ denote the velocity state of the $i$-th pedestrian at time $t$ and 
Without loss of generality, $t = 0$ indicates the current time and $t = -1$ the previous time-step. 
$\mathbf{v}^{i}_{1:T_H}=(\mathbf{v}^i_{1},\dots, \mathbf{v}^i_{T_H})$ represents the future pedestrian velocities over a prediction horizon $T_H$ and $\mathbf{v}^{i}_{-T_O:0}$ the pedestrian past velocities within an observation time $T_O$.
Throughout this paper the subscript $i$ denotes the \emph{query-agent}, i.e., the agent that we want to predict its future motion, and $-i$ the collection of all the other agents. Bold symbols are used to represent vectors and the non bold $x$ and $y$ subscripts are used to refer to the x and y direction in the world frame.  $\mathbf{x}^{i}_0=\{\mathbf{v}^{i}_{-T_O:0},\mathbf{p}^{-i}_{0},\pazocal{O}^{i}_{\textrm{env}}\}$ represents the \textit{query-agent} current state information. %\BF{To shorten the notation we use subscript $\le t$ to represent $t-t_{obs}:t$.}
%Furthermore, we use $t$ to represent the continuous time and $k$ the prediction step index with $k \in [1,T]$. 
To account for the uncertainty and multimodality of the $i$-th pedestrian's motion, we seek a probabilistic model $f(\theta)$ with parameters $\theta$ over a set of $M$ different trajectories $\forall m \in [0,M]$:
%The goal is to find a data-driven model with parameters $\theta$ to estimate the conditional probability of the next $n$ agent states: 
\begin{equation}
    \begin{array}{c}
        p(\mathbf{v}^{i,m}_{1:T_H }| \mathbf{x}^{i}_0) = f( \mathbf{x}^{i}_0,\theta) 
        %\\
        %\forall t \in [1,T]
    \end{array}
\end{equation}
where $m$ is the trajectory index. The probability is conditional to other agents states and the surrounding environment to model the interaction and environment constraints. %For the remaining of this article $\mathbf{x}^{i}_t=\{\mathbf{v}^{i}_{t},\mathbf{p}^{-i}_{t},\pazocal{O}^{i}_{t,\textrm{env}}\}$ represents the model input information.

%$H_t$ defines a hidden state of the model representing its memory of agent's previous states. Therefore, at each time step $t$, the input of our model is composed by current information and the memory state of the model allowing its deployment on a real application. Concerning the multimodal navigation behavior, consider %the dataset not only incorporates the ground truth trajectory%
%that for each agent state $X^i_t$ there is a composite of $M$ different trajectories $X^{i}_{m}$ that the agent may follow. Then, the goal is to find the model parameters $\theta$ which minimize the following loss function:

%which propagates information through time and is updated at every timestep.

%for $ i \in [0 \dots N]$ where N is the number of prediction steps. In this work we assume that the agent velocity follows a bivariate Gaussian distribution $X^{i}_{t} \sim \mathcal{N}(m(x),k(x,x^T))$ and a GP prior over initial agent position $X^{i}_{t} \sim \mathcal{N}(0,0.5) $ and The loss function can now be defined as:

%\begin{equation}
%\theta^* = arg \min_{\theta} \\
%\mathcal{L}^* = \min_{\theta} -\sum^{t_0+T}_{t=t_0} \sum^{M}_{m =0} \log{p(\hat{X}^i_{m,t+1:t+T} | X^{i}_{t},X^{-i}_{t},\pazocal{O}^{\textrm{static}}_{t})} \\
%\end{equation}

%\begin{equation}
%\mathcal{L}^* = & \min_{\theta^*} \sum^{N-1}_{k =t_0} -\log{F(\mathcal{X}_D,\theta)} \\
%\end{equation}

%where $\mathcal{L}$ is the model loss function, $\hat{X}^i_{m,t}$ the predicted $m$-th possible agent state and $t_0$ as the query time instant.

\subsection{Input feature extraction module }\label{sec:inputs}
 This module creates a joint representation of three sources of information: the query-agent state, the environment context and social context. The first input is a sequence of $T_{O}$ history velocities $\mathbf{v}^i_{-T_O:0}$ of the query-agent. %Considering that the agent motion is only governed by its surroundings and not its global position in a map, no position is information is necessary. 
The second input is a local occupancy grid $O^{i}_{\textrm{env}}$, centered at the query-agent containing information about static obstacles (environment context) with width $D_x$ and height $D_y$. Here, we use the global map provided with publicly available datasets \cite{pellegrini2009you,lerner2007crowds}. In a real scenario, the map information can be obtained by building a map offline \cite{zaman2011ros} or local map from online \cite{online_slam} using onboard sensors such as Lidar. %In contrast to \cite{Pfeiffer2017}, we \TBD{TBD}.
%add the surrounding pedestrians to the grid allowing to keep the spacial information about the other pedestrians positions. Still, the network can distinguish between static obstacles and pedestrians, because they are encoded differently.
Due to its high dimensionality, a convolution neural network (CNN) is used to obtain a compressed representation of this occupancy map while maintaining the spatial context. The encoder parameters are obtained by pre-training an Encoder-Decoder structure to minimize $\mathcal{L}_{\textrm{env}}=\sum^{D_x}_{i=1}\sum^{D_y}_{j=1}(\hat{\mathbf{O}^{i}_{\textrm{env}}}-\mathbf{O}^{i}_{\textrm{env}})^2$, as proposed in \cite{Pfeiffer2017}.
%In contrast to \cite{Pfeiffer2017}, we keep the encoder-decoder structure in the VRNN architecture allowing to obtain a reconstructed occupancy grid $\hat{O}^{i}_{\textrm{env}}$. Then, $\hat{O}^{i}_{\textrm{env}}$ is used to continuously train the CNN Encoder-Decoder by minimizing $\mathcal{L}_{\textrm{env}}=\sum^{D_x}_{i=1}\sum^{D_y}_{j=1}(\hat{\mathbf{O}^{i}_{\textrm{env}}}-\mathbf{O}^{i}_{\textrm{env}})^2$ and improve the compressed representation of the static environment, as described in Alg.\ref{alg:grid}. 
In addition, an LSTM layer is added to the first two input channels, modeling the existing time-dependencies. 

The third input provides information about the interaction among the pedestrians containing information about their relative dynamics and spatial configuration. More specifically, it is a vector $\mathbf{O}^{-i}_0=[\mathbf{p}_0^{-1}-\mathbf{p}_0^i,\mathbf{v}_0^{-1}-\mathbf{v}_0^i, \dots \mathbf{p}_0^{-n}-\mathbf{p}_0^i,\mathbf{v}_0^{-n}-\mathbf{v}_0^i]$ with the positions and velocities of the surrounding pedestrians relative to the query-agent. This input vector is then fed into an LSTM, allowing to create a fixed-size representation of the query's agent social context and to consider a variable number of surrounding pedestrians. %Moreover, we add an LSTM layer to the first two input channels to model the existing time-dependencies.
%We propose to model interaction among the pedestrians as a sequence of data, feeding the feature vector $\mathbf{O}^{-i}_0$ into a bidirectional LSTM (BLSTM) \cite{schuster1997bidirectional}. The BLSTM allows to create a fixed-size representation of the query's agent social context and to consider a variable number of surrounding pedestrians. Moreover, it enables information to flow forward and backward
Finally, the outputs of each channel are concatenated creating a compressed and time-dependent representation of the input data
$\mathbf{y}^i=\{\mathbf{y}^{i}_{\mathbf{v}},\mathbf{y}^{i}_{\textrm{env}},\mathbf{y}^{-i}\}$.
Note that we only use past information about the query-agent velocities. For the other inputs only the current information is used.

\subsection{Probabilistic Inference Module}\label{sec:stochectic_module}
The probabilistic inference module is based on the structure of the VRNN, as depicted in Fig. \ref{fig:network_architecture}. It contains three main components: a prior model, a encoder model and decoder model. We use a fully connected layer (FCL) with Relu activation as the encoder model $\psi^{\textrm{enc}}$, the feature extractor of the joint input $\psi^{\mathbf{x}}$ and of the latent random variables $\psi^{\mathbf{z}}$, and the representation of the prior distribution $\psi^{\textrm{prior}}$. $\{\theta_{\textrm{enc}},\theta_{\mathbf{x}},\theta_{\mathbf{z}},\theta_{\textrm{prior}}\}$ are the network parameters of  $\{\psi^{\textrm{enc}},\psi^{\mathbf{x}},\psi^{\mathbf{z}},\psi^{\textrm{prior}}\}$, respectively. The output vectors $\{\psi^{\textrm{enc}}_\tau,\psi^{\textrm{prior}}\}$ are then used to model the approximate posterior  and prior distribution. We split the output vectors into two parts to model the mean and variance, as represented in Fig. \ref{fig:network_architecture}, and apply the following transformations to ensure a valid predicted distribution: $[\mu_{\mathrm{prior}},\mu_{\mathbf{z}}] = [\psi^{\textrm{prior}}_{1:w_{\textrm{prior}}} , \psi^{\textrm{enc}}_{1:w_{\mathbf{z}}}]$ and $[\sigma_{\textrm{prior}},\sigma_{\mathbf{z}}] = [\exp{\psi^{\textrm{prior}}_{w_{\textrm{prior}}:2w_{\textrm{prior}}}},\exp{\psi^{\mathbf{z}}_{w_{\mathbf{z}}:2w_{\mathbf{z}}}}]$.
% $Q(\mathbf{z}_{\le T}|\mathbf{x}_{\le T})$
%As proposed by Graves et. all. \cite{graves2013generating} the following transformations are applied to these output vectors ensuring that the predicted distribution parameters are valid:
\begin{comment}

\begin{equation}
    \label{eq:reparam_sthocastic}
    \begin{split}
          &\mu_{0,t} = \psi^{\textrm{prior}}_{1:w_{\textrm{prior}}} \\
         &\mu_{\mathbf{z},t}  = \psi^{\textrm{enc}}_{1:w_{\mathbf{z}}} \\ 
          &\sigma_{0,t} = \exp{\psi^{\textrm{prior}}_{w_{\textrm{prior}}:2w_{\textrm{prior}}}} \implies \sigma_{0,t} \ge 0  \\
         &\sigma_{\mathbf{z},t} =  \exp{\psi^{\mathbf{z}}_{w_{\mathbf{z}}:2w_{\mathbf{z}}}} \implies \sigma_{0,t} \ge 0 
        %\rho^i_{x,k} = \tanh{\hat{\rho}^i_{x,k}}  \implies \rho^i_{x,k} \in [-1,1] \\
    \end{split}
\end{equation}
\end{comment}
%\begin{subequations}
%\begin{align}
%    \mu_{0,t} = \psi^{prior}_{\tau, 0:w_{prior}} \\
%    \mu_{\mathbf{z},t} = \psi^{enc}_{\tau, 0:w_{\mathbf{z}}} \\
%    \sigma_{0,t} = \textbf{softmax}(\psi^{prior}_{\tau, w_{prior}:2w_{prior}})  \implies \sigma_{0,t} >0 \\
%    \sigma_{\mathbf{z},t} = \textbf{softmax}(\psi^{enc}_{\tau, w_{\mathbf{z}}:2w_{\mathbf{z}}}) \implies \sigma^i_{\mathbf{z},t} >0
    %\rho^i_{x,k} = \tanh{\hat{\rho}^i_{x,k}}  \implies \rho^i_{x,k} \in [-1,1] \\
%\end{align}
%\end{subequations}
$2w_{\textrm{prior}}$ and $2w_{\mathbf{z}}$ are the output vector size of the prior and latent random variable, respectively. This ensures that the standard deviation is always positive. %In contrast to \cite{graves2013generating}, we use a softplus operation to ensure a positive variance.
%Each output tensor is sliced in two parts. To the first is applied a linear layer  modelling the mean $\{\mathbf{\mu}_{enc},\mathbf{\mu}_{prior}\}$ A sofmax operation is applied to the other half of the tensor modelling  $\{\mathbf{\sigma}_{enc},\mathbf{\sigma}_{prior}\}$.
%The first three al to learn the approximate posterior distribution $Q(\mathbf{z}|\mathbf{x})$ as defined in Eq. \ref{eq-posterior}. 
%\REV{From equation (3), it seems like only the last observed timestep of LSTM is fed as input to the decoder, to output all the outputs in one shot i.e mu, sigma from t+1:T. It is not clear whether, at each timestep of LSTM, an output is produced? MAKE CLEAR we use the previous steps and predict multiple steps simultaneously}
 Furthermore, we employ a LSTM layer as the RNN model propagating the hidden-state for the prior model and encoding the time-dependencies for the generative model. In contrast to \cite{vrnn} our generation model conditionally depends on the previous inputs:
\begin{equation}
    \begin{split}
        &\mathbf{\mathbf{v}}_{k} | \mathbf{x}_{0},\mathbf{z}_{0} \sim \mathcal{N}(\mathbf{\mu}_{\mathbf{v},k},\textrm{diag}(\mathbf{\sigma}^2_{\mathbf{v},k})) \\
        &[\mathbf{\mu}_{\mathbf{v},k},\mathbf{\sigma}_{\mathbf{v},k}]=\psi^{\textrm{dec}}(\psi^{\mathbf{z}}(\mathbf{z}_0),\psi^{\mathbf{x}}(\mathbf{y}^i_0),\mathbf{h}_{-1}) \\
        %\forall t \in [1,T]
    \end{split}
\end{equation}
Lastly, the decoder model consists of two FC layers, with ELU \cite{elu} and linear activation, directly connected to the output of the LSTM network. Our models outputs in one shot $T_H$ steps considering the compressed and time-dependent input representation $\mathbf{y}^i_0$.

\subsection{Multi-modal Trajectory Prediction Distribution}\label{sec:output_model}
%\hai{We predict one-shot multi-modal trajectories using our network architecture, in which each modal trajectory is represented by a discrete-time  Gaussian distribution. }

To predict one-shot multi-modal trajectories,%To learn the probability distribution described in Section \ref{sec:problem}, 
 we model the output of our network as a Gaussian Mixture Model (GMM), similar to \cite{bishop1994mixture} and \cite{graves2013generating}, with $M>1$ modes accounting for the multimodality of the pedestrian's motion. For each mode $m \in \{1, \dots, M\}$, we predict a sequence of future pedestrian velocities $v^{i,m}_{1:T_H}$ represented by a bivariate Gaussian $v^{i,m}_k \sim \mathcal{N}(\mu^ {i,m}_{x,k},\mu^{i,m}_{y,k},\sigma^{i,m}_{x,k},\sigma^{i,m}_{y,k}), k = 1, 2, \dots, T_H$, capturing its motion uncertainty. Consequently, a modal trajectory is defined as a sequence of independent bivariate Gaussian's with length $T_H$. The $M$ modes represent a set of $M$ possible trajectories resulting in the following probabilistic model:%with a likelihood of $\pi_m$ to be followed,
\begin{equation}
    p(\mathbf{v}^{i}_{k}|\mathbf{x}_0,\mathbf{z}_{0},\mathbf{h}_{0},\mathbf{\theta}) = \sum_{m=1}^M \pi_m p_G(\mathbf{\mu}^i_{k,m},\mathbf{\sigma}^i_{k,m})
    % \sim  \sum_{m=1}^M\sum_{k=t+1}^{t+T_H}\mathcal{N}(\mathbf{\mu}_{k,m},\mathbf{\sigma}_{k,m}) 
    %p_\theta(\textbf{x}_{t+1:t+T}|\textbf{z}_t)= \varphi_{dec}(\textbf{z}_t,h^{dec}_{t-1}) \\
    %\varphi_{dec}(\textbf{z}_t) =  \pi_m f_{dec}(\textbf{LSTM}(\textbf{concat}(\mathbf{z}_t,\psi_{enc}(\mathbf{x})))   
    %(\mathbf{\mu}_{x,t:t+H},\sigma_{x,t:t+H})
    %\mathbf{\mu}_{x,t:t+H} = \varphi_{enc,k}\\ 
    %\log(\mathbf{\sigma}_{x,t:t+H}^2) = \varphi_{enc,k+w}\\ 
\end{equation}
where $p_G$ is  the probability density function of multivariate Gaussian distributions, $\mathbf{\theta}=\{\theta_{\textrm{enc}},\theta_{\textrm{dec}},\theta_{\mathbf{x}},\theta_{\mathbf{z}},\theta_{\textrm{prior}}\}$ are the model parameters, $\mathbf{\mu}^{i,m}_{k}=[\mathbf{\mu}^{i,m}_{x,k},\mathbf{\mu}^{i,m}_{y,k}]$ and $\mathbf{\sigma}^{i,m}_{k}=[\mathbf{\sigma}^{i,m}_{x,k},\mathbf{\sigma}^{i,m}_{y,k}]$ are the mean and standard deviation of the predicted velocity vectors for the $m$-th predicted trajectory with likelihood $\pi_m$ at time-step $k$, respectively. The transformations described in Sec.\ref{sec:stochectic_module} and Fig.\ref{fig:network_architecture} are applied to the network outputs $\psi^{\textrm{dec}}_{\tau}$ to ensure a valid distribution parametrization.

\subsection{Improving Diversity}\label{sec:sampling}
%\BF{This approach clearly improves the diversity. However, if we follow this approach we loose the likelihood information about following each tracjectory. We could get it, by adding another layer on top to compute those weights but and this time we do not have time for it. This means that we should change our output model.}
Generative models have the key advantage of allowing to perform inference by randomly sampling the latent random variable $\mathbf{z}$ from some prior distribution. Here, we propose a strategy to induce our model to learn a more ``diverse" distribution of trajectories in a interpretable fashion, similar to \cite{rhinehart2018r2p2}. Our VRNN models a generative distribution conditionally dependent on the input representation vector $\mathbf{\mathbf{y}^i}$, which is composed by three sub-vectors $\{\mathbf{y}^{i}_{\mathbf{v}},\mathbf{y}^{i}_{\textrm{env}},\mathbf{y}^{-i}\}$. Now, let's assume that each input vector is a random variable with the following distribution:
%The VRNN models a posterior distribution of $\mathbf{z}$ is conditionally dependent on the input vector $\mathbf{x}$, which is composed by three sub-vectors $\{\mathbf{x}^{i}_{v},\mathbf{x}^{i}_{\textrm{env}},\mathbf{x}^{-i}\}$. Now, let's assume that each input vector is a random variable with the following distribution:
%to condition the prior distribution over $\mathbf{z}$
%\begin{equation}
    %\begin{split}
        %\mathbf{\hat{x}}^k_v = \mathcal{N}(\mathbf{h}_{\mathbf{x}_v},%\sigma_{\mathbf{x}_v}) \\
        %\mathbf{\hat{x}}^k_{env} = \mathcal{N}(\mathbf{h}_{\mathbf{x}_%{env}},\sigma_{\mathbf{x}_{env}})
%    \end{split}
%\end{equation}
\begin{comment}

\begin{equation}
    \begin{split}
        \mathbf{y}^{i}_{\mathbf{v}} &\sim \mathcal{N}(\mathbf{y}^{i}_{0,\mathbf{v}},\sigma_{\mathbf{v}}) \\
        \mathbf{y}^{i}_{\textrm{env}} &\sim \mathcal{N}(\mathbf{y}^{i}_{0,\textrm{env}},\sigma_{\textrm{env}}) \\
        \mathbf{y}^{-i}_{} &\sim \mathcal{N}(\mathbf{y}^{-i}_{0},\sigma_{-i})
    \end{split}
\end{equation}
\end{comment}

\noindent\begin{minipage}{.3\linewidth}
    \begin{equation}
        \mathbf{y}^{i}_{\mathbf{v}} \sim \mathcal{N}(\mathbf{y}^{i}_{0,\mathbf{v}},\sigma_{\mathbf{v}})
    \end{equation}
\end{minipage}
\noindent\begin{minipage}{.3\linewidth}
    \begin{equation}
        \mathbf{y}^{i}_{\textrm{env}} \sim \mathcal{N}(\mathbf{y}^{i}_{0,\textrm{env}},\sigma_{\textrm{env}})
    \end{equation}
\end{minipage}
\noindent\begin{minipage}{.3\linewidth}
    \begin{equation}
        \mathbf{y}^{-i}_{} \sim \mathcal{N}(\mathbf{y}^{-i}_{0},\sigma_{-i})
    \end{equation}
\end{minipage}

%Previous approaches randomly sample $\mathbf{z}$ from a standard Gaussian distribution \cite{alahi2016social,amirian2019social} to generate diverse trajectories. In contrast, we randomly sample from input representation vectors to generate different trajectories. 
where $\{\mathbf{y}^{i}_{\mathbf{v}},\mathbf{y}^{i}_{\textrm{env}},\mathbf{y}^{-i}_{}\}$ are random variables representing the variability of the agent state, the environment and surrounding agents context, respectively. $\{\sigma_{\mathbf{v}},\sigma_{\textrm{env}},\sigma_{-i}\}$ are the variance of each input channel and are considered as hyperparameters of our model. Hence, by sampling from these input distributions we can condition the generation distribution of $\mathbf{x}$ according with the uncertainty on the pedestrian state or the environment context and generate different trajectories $\tilde{\mathbf{v}}^i_{1:T_H}$ by varying the pedestrian conditions. Then, we introduce a loss function which motivates our model to cover the generated trajectories as the following cross-entropy term:

\begin{equation}
    \mathcal{L}_{div}=\sum_{m=1}^M\sum_{k=1}^{T_H}- \mathbb{E}_{}[\log p_G(\mathbf{\tilde{v}}_{k}|\mathbf{x}_{0},\mathbf{z}_{0})] 
\end{equation}

where $\mathbf{\tilde{v}}_{m,k}$ is a velocity sample at time-step $k$ from the m-th sampled trajectory.

\subsection{Training Procedure}\label{sec:train}

The model is trained end-to-end except for the CNN which is pre-trained.
% and trained separately if the reconstruction error is larger than threshold $\epsilon$ but, it has its weights fixed when training overall architecture. 
We train it using back-propagation through time (BTTP) with fixed truncation depth $t_{\textrm{trunc}}$. %For more details about the BTTP process we refer the reader to \cite{Pfeiffer2017}.
%By applying the \textit{reparametrization trick} \cite{kingma2013auto} the latent variable $\mathbf{z}$ is defined as it follows:
Furthermore, we apply the reparametrization trick \cite{kingma2013auto} to obtain a continuous differentiable sampler and train the network using backpropagation. We learn the data distribution by minimizing a timestep-wise variational lower bound with annealing KL-Divergence as loss function \cite{bowman2015generating}:

\begin{subequations}
    \label{eq:recosntruction}
    \begin{align}
        \mathcal{L} = \mathcal{L}_{m} + \lambda *(\mathcal{L}_{\textit{KL}}+\mathcal{L}_{div}) \\
        \mathcal{L}_{m}= \sum_{m=1}^M\sum_{k=1}^{T_H}-\mathbb{E}_{\mathbf{x}_0 \sim \mathbf{D}}[\log \pi_m p_G(\mathbf{v}_{k}|\mathbf{z}_{0},\mathbf{x}_{0})] \\
        \mathcal{L}_{\textit{KL}}(\mathbf{z}_0|\mathbf{x}_{\le 0},\mathbf{z}_{<0}) = \lambda \text{KL}(q(\mathbf{z}_0|\mathbf{x}_{\le 0},\mathbf{z}_{<0})||p_G(\mathbf{z}_0|\mathbf{x}_{< 0},\mathbf{z}_{<0}))
    \end{align}
\end{subequations}
\begin{comment}
\end{comment}
where $\lambda$ is the annealing coefficient.

The first term represents the reconstruction loss (Eq. \ref{eq:recosntruction}b) and the second the KL-Divergence between the approximated posterior $q(\textbf{z}_0|\textbf{x}_{\le 0},\textbf{z}_{<0})$ (Eq. \ref{eq:recosntruction}c) and the prior distribution of $\mathbf{z}$.
%$p(\mathbf{z}_t|\textbf{x}_{< t},\textbf{z}_{<t})$. 
Here, the prior over the latent random variable $\mathbf{z}$ is chosen to be a simple Gaussian distribution with mean and variance $[\mathbf{\mu}_{\textrm{prior},0},\mathbf{\sigma}_{\textrm{prior},0}]=\psi^{\textrm{prior}}(\mathbf{h}_{-1})$ depending on the previous hidden state. During training we aim to find the model parameters which minimize the loss function presented in Equation \ref{eq:recosntruction}a. The annealing coefficient allows the model first to learn the parameters that fit the data well and later in the training phase to match the prior distribution and improve the diversity of the predicted trajectories.%The goal is to find the model parameters $\theta$ which minimize the following loss function:

\section{Experiments}\label{sec:experiments} 

In this section, we show the obtained results of our generative model for simulation and real data. %Firstly, we present the metrics used to evaluate our model. Secondly, detailed settings on the experimental setup used. Thirdly, we demonstrate using a simulated toy dataset that our model can effectively learn different multi-modal trajectories. Moreover, we also perform a qualitative analysis of the predicted trajectories of our model in more complex simulated and real scenarios. Finally, 
We present a qualitative analysis and performance results of our method against three baselines. 
%\subsection{Metrics}
To evaluate the performance of our model against the proposed baselines we use the following evaluation metrics: the average displacement error (ADE) and the final displacement error (FDE). The first two assess the prediction performance. For the models outputting probability distributions, the mean values are used to compute the ADE and FDE metrics. For the multi-modal distributions, we use the trajectory with the minimum error as in \cite{amirian2019social}. %\ja{Good, but you say in the method that we loose the information about likelihood! This is confusing}. \BF{yes if we use the sampling mechanism which I do not use for the performance results. I only use it for the plotted results from real data (Fig4).}
%The NLL metric measures how well the predicted probability distribution fits the observed data and it is used to compare probabilistic models. %%Finally, we employ the mean 2-Wassertein distance to evaluate the variability of the output modes.   
%\begin{equation}
%  \text{M2-W}=\frac{1}{N}\sum_{i=0}^N\sum_{t=0}^T\sum_{m=0}^mW^2_2(P_m,P_{m+1}) 

%\end{equation}
%for $N$ testing sample, $m$ predicted modes with $T$ s of prediction length, and $W^2_2$ as the 2-\textit{Wasserstein} distance. Low values of the M2-W represent similar predicted trajectories for each mode while, high values represent distinct trajectories.
%Thus, this metric allows to evaluate which models collapse to a single mode and which models can effectively predict different trajectories.
%Throughout all the experiments we use the commonly used metrics to evaluate prediction models: the average displacement error (ADE) and the final displacement error (FDE). ADE consists on the average point-wise distance between the real and predicted trajectory, while FDE is just the error between the final predicted and real positions of the query-agent. ADE shows how close the prediction model is from the real navigation behavior. FDE how good the model is predicting the destination of the agent.

\subsection{Experimental Settings}
\vspace{-0.5mm}
We trained our model using RMSProp \cite{tieleman2012rmsprop} which is known to perform well in non-stationary problems with a initial learning rate $\alpha=10^{-4}$ exponentially decaying at a rate of 0.9 and a mini-batch size of 16. We used a KL annealing coefficient $\lambda=\tanh(\frac{\textrm{step}-10^4}{10^3})$, with $\textrm{step}$ as the training step. We set the diversity weight $\beta$ to 0.2 and $\{\sigma_{\mathbf{x}^v},\sigma_{\mathbf{x}^{\textrm{env}}},\sigma_{\mathbf{x}^{-i}}\}=\{0.2,0.2,0\}$. Additionally, to avoid gradient explosion we clip the gradients to 1.0. % and apply a dropout rate of 20\% to the Elu layer of the decoder model. 
We trained and evaluated our model for different prior, latent random variable and input feature vector sizes. The configuration achieving lower validation error was $\{128,128,512\}$, for the prior, latent random variable and input feature vector size, respectively. Moreover, we use $M=3$ mixture components for the models using a GMM as the output function. We set $T_H=12$ prediction steps corresponding to 4.8 s of prediction horizon and $T_O=8$ as used in previous methods \cite{amirian2019social,Sadeghian}. The models were implemented using Tensorflow \cite{tensorflow2015} and were trained on a NVIDIA GeForce GTX 980 requiring $2\times10^4$ training steps, or approximately 2 hours. The simulation datasets were obtained with the open-source ROS implementation of the Social Forces model \cite{Helbing1995}. Our VRNN will be released open source. 

%\subsection{Ablation study}

%\begin{itemize}
%    \item Using a sequence of past velocities improves significantly the performance
%    \item Adding positions did not improve
%    \item depending latent random variable dependent on the previous hidden state improve significantly
%    \item jointly training encoder improves performance
%\end{itemize}
\subsection{Performance evaluation}

We compared our model with the following state-of-art prediction baselines:
\begin{itemize}
    \item \textit{LSTM-D} \cite{Pfeiffer2017}: A deterministic interaction-aware model, incorporating the interaction between the agents and static obstacles.
    \item \textit{SoPhie} \cite{Sadeghian}: a GAN model implementing a Social and Physical attention mechanism.
    \item \textit{Social-ways} (S-Ways) \cite{amirian2019social}: The state-of-art GAN based method for multi-modal trajectory prediction.
    \item \textit{STORN} \cite{bayer2014learning}: Our VRNN model considering a time-independent prior as a Gaussian distribution with zero mean and unit variance.
    %\item \textit{LSTM-MCD} We apply MC-Dropout to the output of the LSTM-D model during inference to predict different trajectories
    %\item \textit{STRON}: We evaluate The STORN model\cite{bayer2014learning} is variant of the VRNN model for wich the prior in independent across the timesteps.
\end{itemize}
\begin{table}[t]
  \caption{Performance results of our proposed method (VRNN) vs. baselines. The results presented for the Social Ways with 30 samples ($K=30$) and SoPhie method were taken from \cite{alahi2016social} and \cite{Sadeghian}, respectively. The ADE and FDE values are separated by slash. The average values (AVG) only consider the results for the real datasets. The results for using three samples ($K=3$) of S-Ways were obtained from the open-source implementation provided by \cite{amirian2019social}.}
         \centering
\begin{tabular}{cccllll}
\hline
\multicolumn{1}{|c}{}                 & \multicolumn{1}{|c|}{Deterministic} & \multicolumn{5}{c|}{Stochastic} \\  \hline
\multicolumn{1}{|c|}{}                 & \multicolumn{1}{c|}{Single Sample} & \multicolumn{2}{c|}{Multiple Samples} & \multicolumn{1}{c|}{} &\multicolumn{2}{c|}{Single Sample} \\  \hline
\multicolumn{1}{|c|}{Dataset}         & \multicolumn{1}{c|}{LSTM}                 & \multicolumn{1}{c|}{SoPhie}   & \multicolumn{1}{c|}{\begin{tabular}[c]{@{}c@{}}S-Ways\\ ($K=30$)\end{tabular}} & \multicolumn{1}{c|}{\begin{tabular}[c]{@{}c@{}}S-Ways\\ ($K=3$)\end{tabular}} & \multicolumn{1}{c|}{STORN}  &   \multicolumn{1}{c|}{VRNN} \\ \hline\hline
%\multicolumn{1}{c||}{\textbf{Simulated}}     &  \multicolumn{1}{c|}{0.37 / 0.81}               &         \multicolumn{1}{c||}{0.36 / 0.37}     &   \multicolumn{1}{c|}{-}    &   \multicolumn{1}{c||}{ - }   &   \multicolumn{1}{c|}{ - }     &    \multicolumn{1}{c|||}{0.66 / 1.27}    &    \multicolumn{1}{c}{\textbf{0.17 / 0.35}}  &    \multicolumn{1}{|c}{ 0.30 / 0.65}  \\  \hline \hline
\multicolumn{1}{|c|}{\textbf{ETH}}               &   \multicolumn{1}{c|}{0.40 / 0.65}     &         \multicolumn{1}{c|}{0.70 / 1.43}  &  \multicolumn{1}{c|}{\textbf{0.39 / 0.64}}  &   \multicolumn{1}{c|}{0.78 / 1.48}        &  \multicolumn{1}{c|}{0.73 / 1.49}     &    \multicolumn{1}{|c|}{\textbf{0.39} / 0.70 }   \\ \hline
\multicolumn{1}{|c|}{\textbf{Hotel}}              &   \multicolumn{1}{c|}{0.45 / 0.75}      &   \multicolumn{1}{c|}{0.76 / 1.67}  &   \multicolumn{1}{c|}{0.39 / 0.64}  &    \multicolumn{1}{c|}{0.53 / 0.95}       &    \multicolumn{1}{c|}{1.33 / 1.45}     &    \multicolumn{1}{|c|}{\textbf{ 0.35 / 0.47 }}   \\ \hline
\multicolumn{1}{|c|}{\textbf{Univ}}              &   \multicolumn{1}{c|}{1.02 / 1.54}     &   \multicolumn{1}{c|}{0.54 / 1.24}  &       \multicolumn{1}{c|}{0.55 / 1.31}       & \multicolumn{1}{c|}{ 0.81 / 1.53 }  &   \multicolumn{1}{c|}{0.82 / 1.17}     &    \multicolumn{1}{c|}{\textbf{0.53 / 0.65}} \\ \hline
\multicolumn{1}{|c|}{\textbf{ZARA01}}           &   \multicolumn{1}{c|}{0.35 / 0.68}     &     \multicolumn{1}{c|}{\textbf{0.30 / 0.63}}  &   \multicolumn{1}{c|}{0.44 / 0.64} &    \multicolumn{1}{c|}{ 0.87 / 1.30 }         &    \multicolumn{1}{c|}{0.91 / 1.52}     &    \multicolumn{1}{|c|}{0.41 / 0.70 }   \\ \hline
\multicolumn{1}{|c|}{\textbf{ZARA02}}           &   \multicolumn{1}{c|}{0.54 / 0.92}     &   \multicolumn{1}{c|}{\textbf{0.38 }/ 0.78}  &   \multicolumn{1}{c|}{0.51 / 0.92} &    \multicolumn{1}{c|}{ 1.27 / 2.13 }         &    \multicolumn{1}{c|}{0.91 / 1.52}    &    \multicolumn{1}{c|}{ 0.51 /\textbf{ 0.55}}  \\ \hline \hline
\multicolumn{1}{|c|}{\textbf{AVG}}           &   \multicolumn{1}{c|}{0.55 / 0.90}     &     \multicolumn{1}{c|}{0.54 / 1.15}  &   \multicolumn{1}{c|}{0.46 / 0.83}   &    \multicolumn{1}{c|}{0.86 / 1.47}       &    \multicolumn{1}{c|}{0.94 / 1.43}   &    \multicolumn{1}{|c|}{\textbf{0.44 / 0.61}}    \\ \hline  \hline
\end{tabular}
\label{tab:global_performance}
\end{table}
%The LSTM-D is a deterministic model, incorporating the interaction between the agents and static obstacles, being used to assess the average prediction performance of our model. The LSTM-MG is a simple expansion of the LSTM-D to model a probability distribution. and Social-ways are two probabilistic models designed for multimodal predictions. The vRNN-NoGrid is used to evaluate the influence of adding the pedestrians to the grid, as described in Alg.\ref{alg:grid}.
%We present experimental result for variants of our model and deterministic baseline: VRNN/LSTM-D is trained with $t_{\textrm{trunc}}=12$ and $T_O=8$ as used in previous methods \cite{amirian2019social,Sadeghian}; VRNN-0/LSTM-0 with $t_{\textrm{trunc}}=1$ and $T_O=0$. 
We use the open-source implementation of \cite{amirian2019social} to obtain the results for S-Ways considering only 3 samples $(K=3)$ as the number of trajectories predicted by our method and as suggested in \cite{trajnet}. We adopt the same dataset split setting as in \cite{amirian2019social} using 4 sets for training and the remaining set for testing.
Aggregated results in Table \ref{tab:global_performance} show that our method outperformed the deterministic baselines, STORN, and S-Ways using three samples. Moreover, the results show that our method achieves comparable performance with state-of-the-art methods using a high number of samples on the Zara01, Zara02 and ETH datasets. In contrast, our method achieves the best performance on the Hotel and Univ datasets. Finally, the poor performance of the STORN model results show that employing a time-dependent prior improves the prediction performance significantly.

\begin{wrapfigure}{r}{.4\textwidth}
  \centering
	\includegraphics[trim=10 10 10 70,width=.4\textwidth]{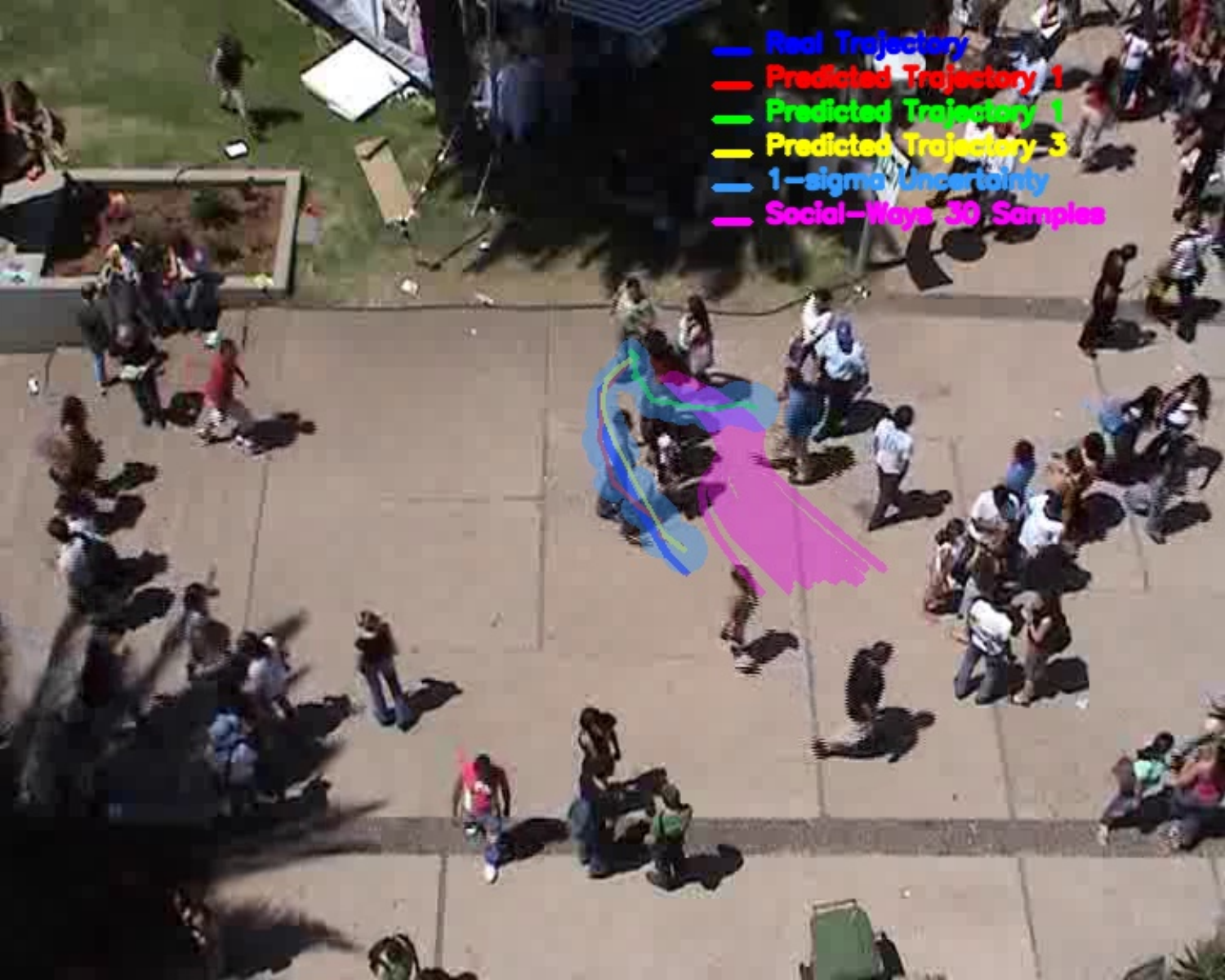}
  \caption{Social-VRNN predicted trajectories vs a multi-modal prediction baseline, Social-Ways \cite{amirian2019social}.
  In blue is depicted the ground truth trajectory, in red, green and yellow the three possible predicted trajectories by our model, in light blue the one sigma boundary of the predicted trajectory and, in magenta 30 sampled predicted trajectories by the Social-Ways model.
  }\label{fig:comparison}
\end{wrapfigure} 

\subsection{Qualitative analysis}
%and, a qualitative evaluation of the predicted trajectories by visual inspection .
\begin{figure*}[!t]
  \centering
  \begin{minipage}{\textwidth}
    \includegraphics[height=3.3cm,width=\textwidth,trim={0cm 0.5cm 0cm 0.5cm},clip]{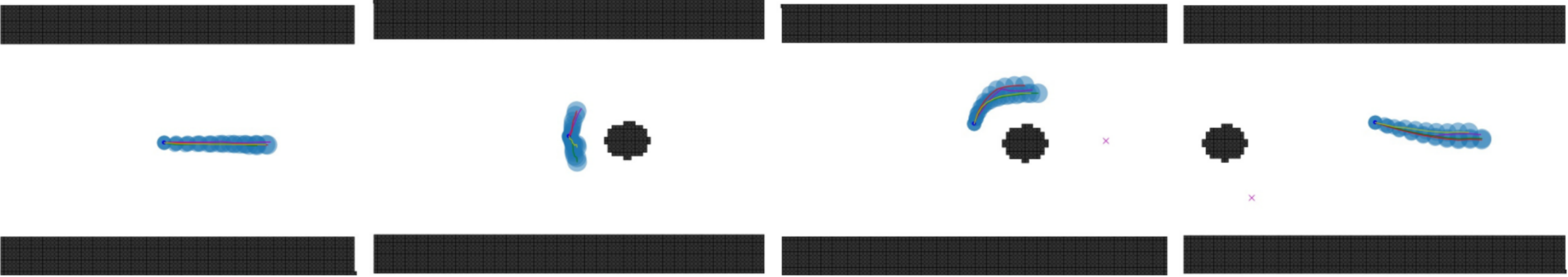} % scene5_step_11
    \caption*{\textit{a)} In this scenario, one agent is moving along a corridor with an obstacle in the middle. The agent is moving from the left to the right. When she finds the obstacle in the middle of its path, our model successfully predicts two hypotheses: going left or right. Once she is already avoiding the obstacle through the left side, the model predicts three hypotheses for the pedestrian to continue its collision avoidance maneuver, with varying clearance levels. Finally, when she is in free space all the predicted trajectories collapse to a single-mode.}
    \label{fig:test1} 
  \end{minipage}
    \hspace{10mm}
    \begin{minipage}{\textwidth}
      \includegraphics[height=3.3cm,width=\textwidth,trim={0cm 0cm 0cm 0cm},clip]{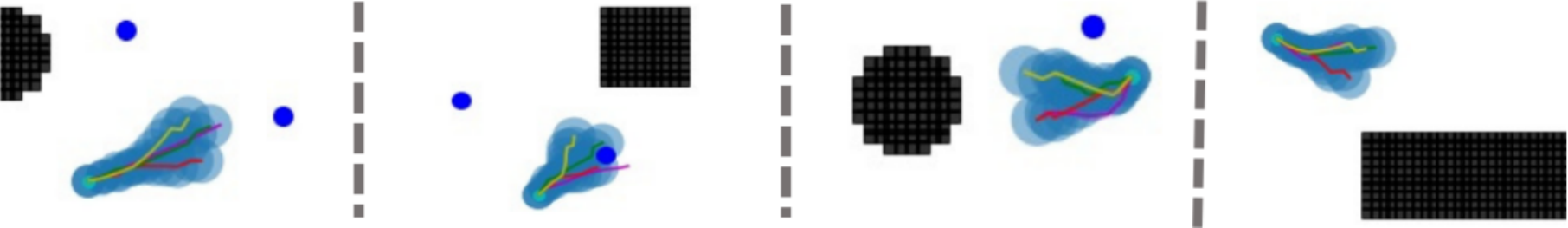} %scene12_step_22 scene12_step_20 is a good one 13_8 too
      \caption*{\textit{b)} This sub-figure illustrates four sample  results obtained in a more complex simulated scenario, with several static obstacles and 15 agents. The two left figures show two situations where the agent can avoid another agent on its left, right or by simply move straight because the other will keep moving away. The two right figures show the ability of our model to predict different trajectories that an agent may follow to avoid a static obstacle.}
      \label{fig:test2}
    \end{minipage}
    \hspace{10mm}
    \begin{minipage}{\textwidth}
      \includegraphics[height=3.3cm,width=\textwidth,trim={0cm 0.5cm 0cm 0.5cm},clip]{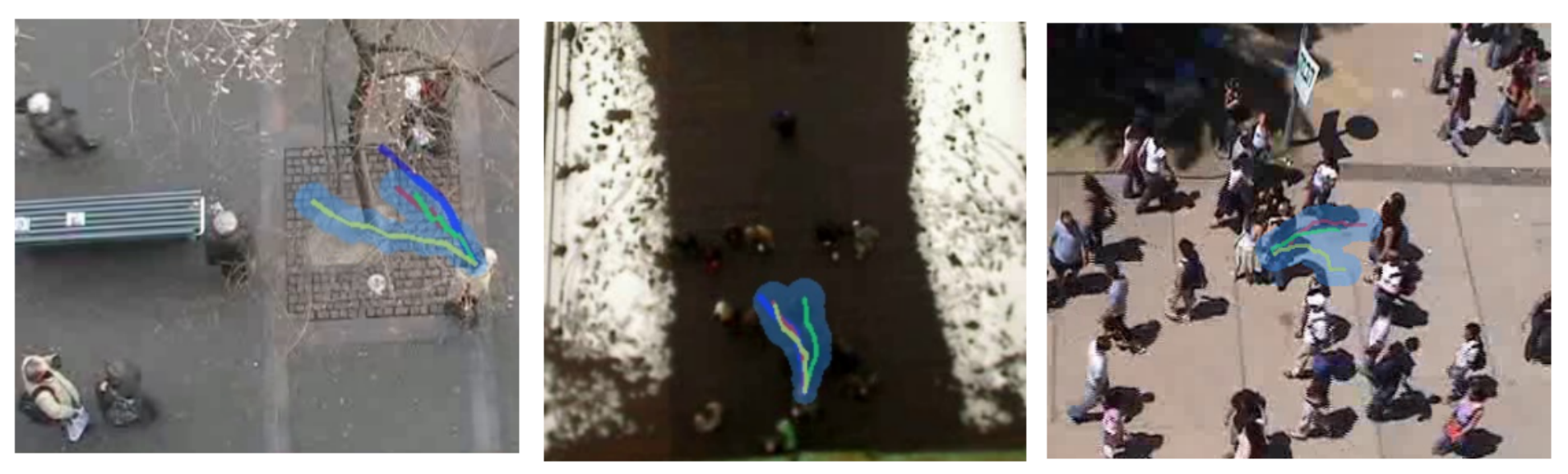} %scene12_step_22 scene12_step_20 is a good one 13_8 too / result_15_22_zoom
      \caption*{\textit{c)} Three examples of multi-modal trajectory prediction using our model in real scenarios. In blue is depicted the ground truth trajectory, in red, green and yellow the three possible predicted trajectories, in light blue the one sigma boundary of the predicted trajectory.}
      \label{fig:real_scenarios}
    \end{minipage}
  \caption{ The scenarios depicted in Fig.\ref{fig:toy_experient}(a) and (b) were simulated by using the Social Forces model \cite{Helbing1995} for the pedestrians. In magenta the real trajectory, in red, green and yellow the mean values of each trajectory hypothesis and, in blue the 1-$\sigma$ uncertainty boundaries of each trajectory. The dark blue dots represent the other agents. The plotted trajectories correspond to a single network query.%to our network model. %\ja{Do we really need to show the whole map? If we zoom in to the area of interest the predictions and obstacles will be much clearer}
  }
  \label{fig:toy_experient}
\end{figure*}  

In this section we present prediction results for simulated and real scenarios, as depicted in Fig. \ref{fig:toy_experient}.
We have created two datasets to demonstrate this multi-modal behavior with static obstacles (Fig.\ref{fig:toy_experient}(a)), and other pedestrians (Fig. \ref{fig:toy_experient}(b)). Figure \ref{fig:toy_experient}(a) shows the ability of our method to predict different trajectories according to the environment structure. Figure \ref{fig:toy_experient}(b) demonstrates that our method can scale to more complex environments, with several pedestrians and obstacles, and predict different motion hypotheses. %Note that we use only a single sample for the depicted results.
%Firstly, we start by analyzing two simple scenarios to demonstrate the ability of our generative model to learn multiple hypotheses. Then, we evaluate our model on a complex simulated scenario with several obstacles and pedestrians, and on real data. 

%In the second scenario, Fig.\ref{fig:toy_experient}(d) to Fig.\ref{fig:toy_experient}(f), four agent's walk back and forward along a free corridor. Fig.\ref{fig:toy_experient}(d) and Fig.\ref{fig:toy_experient}(e) present two specific situation when a pedestrian walking from left to right finds other pedestrians walking in opposite direction. In Fig.\ref{fig:toy_experient}(f), the pedestrian keeps walking to the right side and because there are any pedestrian crossing with she all the predict trajectories collapse to a single one.s
Moreover, we evaluate our method on real data using the publicly available datasets \cite{pellegrini2009you,lerner2007crowds}. % in two scenarios: a hotel reception (Fig. \ref{fig:real_scenarios}(a)) and a faculty entrance lobby (Fig. \ref{fig:real_scenarios}(b)). % Notwithstanding the lask of interaction situations in this dataset
%Predicting multimodal trajectories is extremely challenging in these two scenarios, because these datasets lack situations of interaction between the pedestrians and static obstacles. Yet, as depicted in Fig. \ref{fig:real_scenarios} our method successfully different modal trajectories.
%As stated in \cite{amirian2019social}, predicting multi-modal trajectories is extremely difficult because the existing publicly available datasets lack situations of interaction between the pedestrians and static obstacles. Thus, for the real results we use the sampling mechanism proposed Section \ref{sec:sampling}. We use three samples with $\{\sigma_{\mathbf{x}^v},\sigma_{\mathbf{x}^{\textrm{env}}},\sigma_{\mathbf{x}^{-i}}\}=\{0.5,0,0\}$. Adding some stochasticity to the agent state representation $\mathbf{y}_v$ was sufficient for the model to generate diverse trajectories.
In Fig. \ref{fig:toy_experient}(c) on the left, our model infers two possible trajectories for the pedestrian to avoid a tree. In addition, in the central and right images of  Fig. \ref{fig:toy_experient}(c), our model predicts two possible trajectories to move through the crowd. Finally, Fig. \ref{fig:comparison} shows predicted trajectories for both Social-VRNN and Social-Ways model in a crowded scene. The predicted trajectories from the Social-VRNN model can capture two distinct trajectories through the crowd. In contrast, Social-Ways only captures one mode, even considering 30 samples from the baseline model. The presented results demonstrate that our model can effectively infer different trajectories according to the environment and social constraints from a single query. We refer the reader to the video\footnote{\url{https://youtu.be/tBr5v7TXyG0}} accompanying this article for more details on the presented results.

\section{CONCLUSION}\label{sec:conclusion}

%The incorporation of the agents on the static grid map allows a reduced number of predicted trajectories which may result in collision.

%2-Wasserstein looks to generate more different trajectories or better average w2 score.

%Adding agents on the grid improves significantly the performance.
In this paper, we introduced a Variational Recurrent Neural Network (VRNN) architecture for multi-modal trajectory prediction in one-shot and considering the pedestrian dynamics, interactions among pedestrians and static obstacles. Building on a variational approach and
%significantly reduce the number of training iterations, as well as the number of samples to achieve state-of-art performance from a single network query. 
learning a mixture Gaussian model enables our model to generate distinct trajectories accounting for the static obstacles and the surrounding pedestrians. Our approach allows us to improve the state-of-the-art prediction performance in scenarios with a large number of agents (e.g., Univ dataset) or containing static obstacles (e.g., Hotel dataset) from a single prediction shot. Furthermore, the proposed approach reduces significantly the number of samples needed to achieve good prediction with high accuracy.
%Finally, we have also proposed a training strategy to improve the diversity of the predicted trajectories. We propose to sample diverse trajectories from our generative model. Then, during training we minimize a loss function to fit both trajectories from data or sampled from the model.
As future work, we aim %to expand the proposed model to learn the likelihood separately allowing to use the sampling mechanism without losing this information. Moreover, we intend 
to integrate the proposed method with a real-time motion planner on a mobile platform for autonomous navigation among pedestrians.

% The acknowledgments are automatically included only in the final version of the paper.
\acknowledgments{This work was supported by the Amsterdam Institute for Advanced Metropolitan Solutions and the Netherlands Organisation for Scientific Research (NWO) domain Applied Sciences (Veni 15916).}

%===============================================================================

\bibliography{main}  % .bib

\begin{thebibliography}{41}
\providecommand{\natexlab}[1]{#1}
\providecommand{\url}[1]{\texttt{#1}}
\expandafter\ifx\csname urlstyle\endcsname\relax
  \providecommand{\doi}[1]{doi: #1}\else
  \providecommand{\doi}{doi: \begingroup \urlstyle{rm}\Url}\fi

\bibitem[{Brito} et~al.(2019){Brito}, {Floor}, {Ferranti}, and
  {Alonso-Mora}]{8768044}
B.~{Brito}, B.~{Floor}, L.~{Ferranti}, and J.~{Alonso-Mora}.
\newblock Model predictive contouring control for collision avoidance in
  unstructured dynamic environments.
\newblock \emph{IEEE Robotics and Automation Letters}, 4\penalty0 (4):\penalty0
  4459--4466, 2019.

\bibitem[Lin et~al.(2020)Lin, Zhu, and Alonso-Mora]{Zhu2020ICRA}
J.~Lin, H.~Zhu, and J.~Alonso-Mora.
\newblock {Robust vision-based obstacle avoidance for micro aerial vehicles in
  dynamic environments}.
\newblock In \emph{2020 IEEE International Conference on Robotics and
  Automation (ICRA)}, pages 2682--2688. IEEE, 2020.

\bibitem[Kothari et~al.(2020)Kothari, Kreiss, and Alahi]{kothari2020human}
P.~Kothari, S.~Kreiss, and A.~Alahi.
\newblock Human trajectory forecasting in crowds: A deep learning perspective.
\newblock \emph{arXiv preprint arXiv:2007.03639}, 2020.

\bibitem[Pfeiffer et~al.(2018)Pfeiffer, Paolo, Sommer, Nieto, Siegwart, and
  Cadena]{Pfeiffer2017}
M.~Pfeiffer, G.~Paolo, H.~Sommer, J.~Nieto, R.~Siegwart, and C.~Cadena.
\newblock A data-driven model for interaction-aware pedestrian motion
  prediction in object cluttered environments.
\newblock In \emph{2018 IEEE International Conference on Robotics and
  Automation (ICRA)}, pages 1--8. IEEE, 2018.

\bibitem[L{\"o}tjens et~al.(2019)L{\"o}tjens, Everett, and
  How]{lotjens2019safe}
B.~L{\"o}tjens, M.~Everett, and J.~P. How.
\newblock Safe reinforcement learning with model uncertainty estimates.
\newblock In \emph{2019 International Conference on Robotics and Automation
  (ICRA)}, pages 8662--8668. IEEE, 2019.

\bibitem[Gal and Ghahramani(2016)]{gal2016dropout}
Y.~Gal and Z.~Ghahramani.
\newblock Dropout as a bayesian approximation: Representing model uncertainty
  in deep learning.
\newblock In \emph{international conference on machine learning}, pages
  1050--1059, 2016.

\bibitem[Gupta et~al.(2018)Gupta, Johnson, Fei-Fei, Savarese, and
  Alahi]{Gupta2018}
A.~Gupta, J.~Johnson, L.~Fei-Fei, S.~Savarese, and A.~Alahi.
\newblock Social gan: Socially acceptable trajectories with generative
  adversarial networks.
\newblock In \emph{Proceedings of the IEEE Conference on Computer Vision and
  Pattern Recognition}, pages 2255--2264, 2018.

\bibitem[Amirian et~al.(2019)Amirian, Hayet, and Pettr{\'e}]{amirian2019social}
J.~Amirian, J.-B. Hayet, and J.~Pettr{\'e}.
\newblock Social ways: Learning multi-modal distributions of pedestrian
  trajectories with gans.
\newblock In \emph{Proceedings of the IEEE Conference on Computer Vision and
  Pattern Recognition Workshops}, pages 0--0, 2019.

\bibitem[Chung et~al.(2015)Chung, Kastner, Dinh, Goel, Courville, and
  Bengio]{vrnn}
J.~Chung, K.~Kastner, L.~Dinh, K.~Goel, A.~C. Courville, and Y.~Bengio.
\newblock A recurrent latent variable model for sequential data.
\newblock In \emph{Advances in neural information processing systems}, pages
  2980--2988, 2015.

\bibitem[Helbing and Molnar(1995)]{Helbing1995}
D.~Helbing and P.~Molnar.
\newblock {Social force model for pedestrian dynamics}, 1995.
\newblock ISSN 1063651X.

\bibitem[Kim et~al.(2015)Kim, Guy, Liu, Wilkie, Lau, Lin, and Manocha]{Kim2015}
S.~Kim, S.~J. Guy, W.~Liu, D.~Wilkie, R.~W. Lau, M.~C. Lin, and D.~Manocha.
\newblock {BRVO: Predicting pedestrian trajectories using velocity-space
  reasoning}.
\newblock \emph{International Journal of Robotics Research}, 34\penalty0
  (2):\penalty0 201--217, 2015.
\newblock ISSN 17413176.

\bibitem[Trautman and Krause(2010)]{Trautman2010}
P.~Trautman and A.~Krause.
\newblock {Unfreezing the robot: Navigation in dense, interacting crowds}.
\newblock In \emph{IEEE/RSJ 2010 International Conference on Intelligent Robots
  and Systems, IROS 2010 - Conference Proceedings}, pages 797--803, 2010.
\newblock ISBN 9781424466757.

\bibitem[Becker et~al.(2018)Becker, Hug, and Arens]{Becker2018}
S.~Becker, R.~Hug, and M.~Arens.
\newblock {An Evaluation of Trajectory Prediction Approaches and Notes on the
  TrajNet Benchmark.}
\newblock 2018.

\bibitem[Xue et~al.(2018)Xue, Huynh, and Reynolds]{Xue2018}
H.~Xue, D.~Q. Huynh, and M.~Reynolds.
\newblock Ss-lstm: A hierarchical lstm model for pedestrian trajectory
  prediction.
\newblock In \emph{2018 IEEE Winter Conference on Applications of Computer
  Vision (WACV)}, pages 1186--1194. IEEE, 2018.

\bibitem[Alahi et~al.(2016)Alahi, Goel, Ramanathan, Robicquet, Fei-Fei, and
  Savarese]{alahi2016social}
A.~Alahi, K.~Goel, V.~Ramanathan, A.~Robicquet, L.~Fei-Fei, and S.~Savarese.
\newblock Social lstm: Human trajectory prediction in crowded spaces.
\newblock In \emph{Proceedings of the IEEE Conference on Computer Vision and
  Pattern Recognition}, pages 961--971, 2016.

\bibitem[Hasan et~al.(2018)Hasan, Setti, Tsesmelis, Del~Bue, Galasso, and
  Cristani]{hasan2018mx}
I.~Hasan, F.~Setti, T.~Tsesmelis, A.~Del~Bue, F.~Galasso, and M.~Cristani.
\newblock Mx-lstm: mixing tracklets and vislets to jointly forecast
  trajectories and head poses.
\newblock In \emph{Proceedings of the IEEE Conference on Computer Vision and
  Pattern Recognition}, pages 6067--6076, 2018.

\bibitem[Bartoli et~al.(2018)Bartoli, Lisanti, Ballan, and
  Del~Bimbo]{bartoli2018context}
F.~Bartoli, G.~Lisanti, L.~Ballan, and A.~Del~Bimbo.
\newblock Context-aware trajectory prediction.
\newblock In \emph{2018 24th International Conference on Pattern Recognition
  (ICPR)}, pages 1941--1946. IEEE, 2018.

\bibitem[Xu et~al.(2018)Xu, Qin, Wang, Huang, Ye, and Zhang]{xu2018collision}
K.~Xu, Z.~Qin, G.~Wang, K.~Huang, S.~Ye, and H.~Zhang.
\newblock Collision-free lstm for human trajectory prediction.
\newblock In \emph{International Conference on Multimedia Modeling}, pages
  106--116. Springer, 2018.

\bibitem[Lerner et~al.(2007)Lerner, Chrysanthou, and
  Lischinski]{lerner2007crowds}
A.~Lerner, Y.~Chrysanthou, and D.~Lischinski.
\newblock Crowds by example.
\newblock In \emph{Computer graphics forum}, volume~26, pages 655--664. Wiley
  Online Library, 2007.

\bibitem[Sadeghian et~al.(2019)Sadeghian, Kosaraju, Sadeghian, Hirose,
  Rezatofighi, and Savarese]{Sadeghian}
A.~Sadeghian, V.~Kosaraju, A.~Sadeghian, N.~Hirose, H.~Rezatofighi, and
  S.~Savarese.
\newblock Sophie: An attentive gan for predicting paths compliant to social and
  physical constraints.
\newblock In \emph{Proceedings of the IEEE Conference on Computer Vision and
  Pattern Recognition}, pages 1349--1358, 2019.

\bibitem[Makansi et~al.(2019)Makansi, Ilg, Cicek, and
  Brox]{makansi2019overcoming}
O.~Makansi, E.~Ilg, O.~Cicek, and T.~Brox.
\newblock Overcoming limitations of mixture density networks: A sampling and
  fitting framework for multimodal future prediction.
\newblock In \emph{Proceedings of the IEEE Conference on Computer Vision and
  Pattern Recognition}, pages 7144--7153, 2019.

\bibitem[Zhao et~al.(2019)Zhao, Xu, Monfort, Choi, Baker, Zhao, Wang, and
  Wu]{zhao2019multi}
T.~Zhao, Y.~Xu, M.~Monfort, W.~Choi, C.~Baker, Y.~Zhao, Y.~Wang, and Y.~N. Wu.
\newblock Multi-agent tensor fusion for contextual trajectory prediction.
\newblock In \emph{Proceedings of the IEEE Conference on Computer Vision and
  Pattern Recognition}, pages 12126--12134, 2019.

\bibitem[Chandra et~al.(2020)Chandra, Guan, Panuganti, Mittal, Bhattacharya,
  Bera, and Manocha]{chandra2020forecasting}
R.~Chandra, T.~Guan, S.~Panuganti, T.~Mittal, U.~Bhattacharya, A.~Bera, and
  D.~Manocha.
\newblock Forecasting trajectory and behavior of road-agents using spectral
  clustering in graph-lstms.
\newblock \emph{IEEE Robotics and Automation Letters}, 5\penalty0 (3):\penalty0
  4882--4890, 2020.

\bibitem[Eiffert et~al.(2020)Eiffert, Li, Shan, Worrall, Sukkarieh, and
  Nebot]{eiffert2020probabilistic}
S.~Eiffert, K.~Li, M.~Shan, S.~Worrall, S.~Sukkarieh, and E.~Nebot.
\newblock Probabilistic crowd gan: Multimodal pedestrian trajectory prediction
  using a graph vehicle-pedestrian attention network.
\newblock \emph{IEEE Robotics and Automation Letters}, 5\penalty0 (4):\penalty0
  5026--5033, 2020.

\bibitem[Salzmann et~al.(2020)Salzmann, Ivanovic, Chakravarty, and
  Pavone]{salzmann2020trajectron++}
T.~Salzmann, B.~Ivanovic, P.~Chakravarty, and M.~Pavone.
\newblock Trajectron++: Dynamically-feasible trajectory forecasting with
  heterogeneous data.
\newblock \emph{arXiv preprint arXiv:2001.03093}, 2020.

\bibitem[Zhi et~al.(2019)Zhi, Senanayake, Ott, and
  Ramos]{zhi2019spatiotemporal}
W.~Zhi, R.~Senanayake, L.~Ott, and F.~Ramos.
\newblock Spatiotemporal learning of directional uncertainty in urban
  environments with kernel recurrent mixture density networks.
\newblock \emph{IEEE Robotics and Automation Letters}, 4\penalty0 (4):\penalty0
  4306--4313, 2019.

\bibitem[Zhi et~al.(2020)Zhi, Ott, and Ramos]{zhi2020kernel}
W.~Zhi, L.~Ott, and F.~Ramos.
\newblock Kernel trajectory maps for multi-modal probabilistic motion
  prediction.
\newblock In \emph{Conference on Robot Learning}, pages 1405--1414, 2020.

\bibitem[Ivanovic and Pavone(2019)]{trajectron}
B.~Ivanovic and M.~Pavone.
\newblock The trajectron: Probabilistic multi-agent trajectory modeling with
  dynamic spatiotemporal graphs.
\newblock pages 2375--2384, 10 2019.

\bibitem[Pellegrini et~al.(2009)Pellegrini, Ess, Schindler, and
  Van~Gool]{pellegrini2009you}
S.~Pellegrini, A.~Ess, K.~Schindler, and L.~Van~Gool.
\newblock You'll never walk alone: Modeling social behavior for multi-target
  tracking.
\newblock In \emph{2009 IEEE 12th International Conference on Computer Vision},
  pages 261--268. IEEE, 2009.

\bibitem[Zaman et~al.(2011)Zaman, Slany, and Steinbauer]{zaman2011ros}
S.~Zaman, W.~Slany, and G.~Steinbauer.
\newblock Ros-based mapping, localization and autonomous navigation using a
  pioneer 3-dx robot and their relevant issues.
\newblock In \emph{2011 Saudi International Electronics, Communications and
  Photonics Conference (SIECPC)}, pages 1--5. IEEE, 2011.

\bibitem[{Huang} et~al.(2005){Huang}, {Rad}, and {Wong}]{online_slam}
G.~Q. {Huang}, A.~B. {Rad}, and Y.~K. {Wong}.
\newblock Online slam in dynamic environments.
\newblock In \emph{ICAR '05. Proceedings., 12th International Conference on
  Advanced Robotics, 2005.}, pages 262--267, 2005.

\bibitem[Clevert et~al.(2015)Clevert, Unterthiner, and Hochreiter]{elu}
D.-A. Clevert, T.~Unterthiner, and S.~Hochreiter.
\newblock Fast and accurate deep network learning by exponential linear units
  (elus).
\newblock \emph{CoRR}, abs/1511.07289, 2015.

\bibitem[Bishop(1994)]{bishop1994mixture}
C.~M. Bishop.
\newblock Mixture density networks.
\newblock Technical report, Citeseer, 1994.

\bibitem[Graves(2013)]{graves2013generating}
A.~Graves.
\newblock Generating sequences with recurrent neural networks, 2013.

\bibitem[Rhinehart et~al.(2018)Rhinehart, Kitani, and
  Vernaza]{rhinehart2018r2p2}
N.~Rhinehart, K.~M. Kitani, and P.~Vernaza.
\newblock R2p2: A reparameterized pushforward policy for diverse, precise
  generative path forecasting.
\newblock In \emph{Proceedings of the European Conference on Computer Vision
  (ECCV)}, pages 772--788, 2018.

\bibitem[Kingma and Welling(2013)]{kingma2013auto}
D.~P. Kingma and M.~Welling.
\newblock Auto-encoding variational bayes.
\newblock \emph{CoRR}, abs/1312.6114, 2013.

\bibitem[Bowman et~al.(2016)Bowman, Vilnis, Vinyals, Dai, Jozefowicz, and
  Bengio]{bowman2015generating}
S.~R. Bowman, L.~Vilnis, O.~Vinyals, A.~Dai, R.~Jozefowicz, and S.~Bengio.
\newblock Generating sentences from a continuous space.
\newblock \emph{Proceedings of The 20th SIGNLL Conference on Computational
  Natural Language Learning}, 2016.
\newblock URL \url{http://dx.doi.org/10.18653/v1/k16-1002}.

\bibitem[Tieleman and Hinton(2012)]{tieleman2012rmsprop}
T.~Tieleman and G.~Hinton.
\newblock Rmsprop: Divide the gradient by a running average of its recent
  magnitude. coursera: Neural networks for machine learning.
\newblock \emph{Tech. Rep., Technical report}, page~31, 2012.

\bibitem[Abadi et~al.(2015)Abadi, Agarwal, Barham, Brevdo, Chen, Citro,
  Corrado, Davis, Dean, Devin, et~al.]{tensorflow2015}
M.~Abadi, A.~Agarwal, P.~Barham, E.~Brevdo, Z.~Chen, C.~Citro, G.~S. Corrado,
  A.~Davis, J.~Dean, M.~Devin, et~al.
\newblock Tensorflow: Large-scale machine learning on heterogeneous systems,
  2015.
\newblock \emph{Software available from tensorflow. org}, 1\penalty0 (2), 2015.

\bibitem[Bayer and Osendorfer(2014)]{bayer2014learning}
J.~Bayer and C.~Osendorfer.
\newblock Learning stochastic recurrent networks.
\newblock \emph{arXiv preprint arXiv:1411.7610}, 2014.

\bibitem[tra()]{trajnet}
{Trajnet++ (A Trajectory Forecasting Challenge)}.
\newblock URL
  \url{https://www.aicrowd.com/challenges/trajnet-a-trajectory-forecasting-challenge}.

\end{thebibliography}

\end{document}